\documentclass[letterpaper, 10pt, conference]{ieeeconf}
\usepackage{graphicx}
\graphicspath{{images/}}
\usepackage{multirow}
\usepackage{float}
\usepackage{color}
\usepackage{tabularx}
\usepackage{amsmath}
\usepackage[colorlinks=true, urlcolor=blue, linkcolor=blue]{hyperref}
\usepackage{cite}
\usepackage[font=small,tableposition=top,hypcap=false]{caption}

\IEEEoverridecommandlockouts
\title{\LARGE \bf Robust SLAM Systems: Are We There Yet?
}

\author{Mihai Bujanca$^1$, Xuesong Shi$^2$, Matthew Spear$^1$, Pengpeng Zhao$^{2,3}$, Barry Lennox$^1$, Mikel Luj\'an$^1$%
\thanks{$^{1}$ The University of Manchester, Manchester, UK}%
\thanks{$^{2}$ Intel Labs China, Beijing, China}%
\thanks{$^{3}$ Beihang University, Beijing, China}}

\makeatletter
\let\@oldmaketitle\@maketitle
\renewcommand{\@maketitle}{\@oldmaketitle
\centering\includegraphics[width=0.99\textwidth]{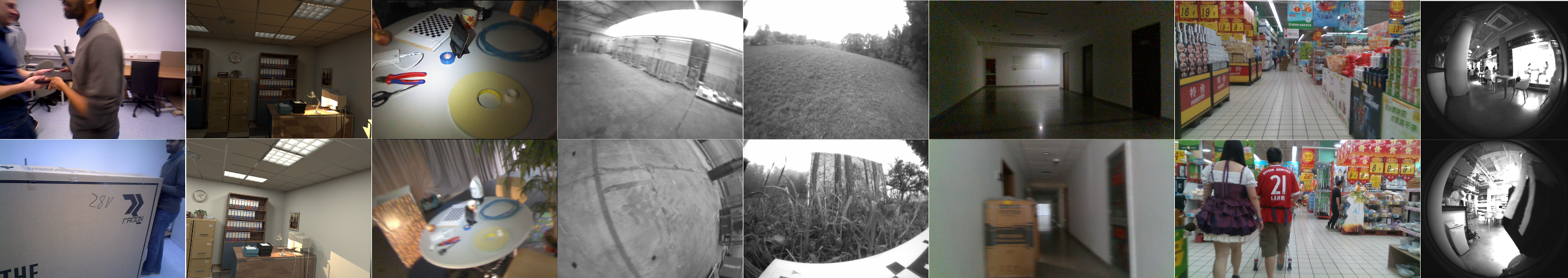}\\
\captionof{figure}{Sample frames exhibiting challenging perturbations; \textit{e.g.}\ occlusions and dynamically-moving elements; frames from drone sequences containing motion blur and no reliable features; frames containing lighting differences, blur, and dynamic objects.}
\label{fig:samples}
 }
 \let\NAT@parse\undefined
\makeatother

\begin{document}
\maketitle
\begin{abstract}
Progress in the last decade has brought about significant improvements in the accuracy and speed of SLAM systems, broadening their mapping capabilities. Despite these advancements, long-term operation remains a major challenge, primarily due to the wide spectrum of perturbations robotic systems may encounter.

Increasing the robustness of SLAM algorithms is an ongoing effort, however it usually addresses a specific perturbation. Generalisation of robustness across a large variety of challenging scenarios is not well-studied nor understood. This paper presents a systematic evaluation of the robustness of open-source state-of-the-art SLAM algorithms with respect to challenging conditions such as fast motion, non-uniform illumination, and dynamic scenes. The experiments are performed with perturbations present both independently of each other, as well as in combination in long-term deployment settings in unconstrained environments (\textit{lifelong operation}). 

The detailed results (approx.\ 20,000 experiments) along with comprehensive documentation of the benchmarking tool for integrating new datasets and evaluating SLAM algorithms not studied in this work are available at \href{https://robustslam.github.io/evaluation}{https://robustslam.github.io/evaluation}.
\end{abstract}

\begin{table*}[tbh!]
\begin{center}
    \begin{tabular}{ | l | l | l | l | l |}
    \hline
    Name                                    & Sensors               & Perturbations                             & Platform              & Year      \\ \hline
                                                                                                                                                                \hline
    OpenLORIS \cite{shi2020we}              & RGB-D, Stereo, LiDAR  & Dynamic movement, sensor degradation,             & Ground robot          & 2020   \\  
                                            & IMU, Wheel odometry   & changed viewpoints and objects, illumination                &                       &        \\
    BONN Dynamic \cite{palazzolo2019refusion}  & RGB-D                 & Dynamic movement                                  & Handheld              & 2019   \\
    ETHI \cite{park2017icra}                & RGB-D                 & Illumination changes                              & Synthetic, handheld   & 2017               \\ 
    EuRoC MAV \cite{Burri25012016}          & Stereo, IMU           & Baseline dataset; Motion blur                     & Drone & 2016         \\ 
    ICL-NUIM \cite{iclnuim}                 & RGB-D                 & Baseline dataset                                  & Synthetic             & 2014     \\ 
    TUM RGB-D \cite{sturm12iros}            & RGB-D                 & Baseline dataset; Dynamic objects,    & Handheld              & 2012    \\ \hline
    \end{tabular}
    \caption{Datasets used in the evaluation.}
\label{tab:slambench_datasets}
\end{center}
\end{table*}

\begin{table*}[tbh!]
\centering
\vspace{-3em}
\begin{tabular}{|l|l|l|l|l|}
\hline
    Algorithm                                           & Type                   & Sensors                      & Processing             & Year \\\hline
                                                                                                                                                 \hline
    OpenVINS \cite{Geneva2020ICRA}                      & Sparse                 & Stereo, IMU                  & CPU                     & 2020   \\
    ORB-SLAM3 \cite{ORBSLAM3_2020}                      & Sparse                 & RGB-D, Stereo, Monocular, IMU & CPU                  & 2020   \\
    FullFusion \cite{bujanca2019fullfusion}             & Dense, non-rigid, semantic     & RGB-D                        & GPU                     & 2019   \\
    ReFusion \cite{palazzolo2019refusion}               & Dense                  & RGB-D                        & GPU                  & 2019   \\
    ORB-SLAM2 \cite{murORB2}                            & Sparse                 & RGB-D, Stereo, Monocular     & CPU                    & 2016   \\
    ElasticFusion \cite{whelan2015elasticfusion}        & Dense                  & RGB-D                        & GPU                    & 2015   \\
    \hline
    \end{tabular}
    \caption{SLAM systems evaluated.}
    \label{tab:slam_algorithms}
\vspace{-2em}
\end{table*}

\section{Introduction}
SLAM algorithms are an essential component of embodied AI systems, providing a fundamental infrastructure necessary for navigation and other high-level tasks. 
The progress of SLAM systems during the last three decades has been remarkable, improving both the localisation and the mapping capabilities. While initially only very small spaces such as table tops or small rooms could be mapped, today's SLAM algorithms can operate on large scales \cite{bosse2004simultaneous,lynen2015get}. Thanks to advancements in computing hardware, sensors, and machine learning, SLAM has also extended well beyond the initial landmark-based mapping, leading to dense 3D reconstruction, non-rigid 3D reconstruction, and semantic mapping. The localisation accuracy of SLAM systems has also improved dramatically: the top 40 submissions on the KITTI odometry benchmark \cite{Geiger2012CVPR} have errors below 1\%. 

Thanks to these advancements, SLAM has enabled new applications and while opportunities for further improvement remain ahead, robustness is widely regarded as today's most difficult challenge \cite{Cadena16tro-SLAMfuture}. We define \textit{robustness} as the capacity of a system to avoid fatal failures either by continuously performing accurately, or by detecting and quickly recovering from soft failures. A \textit{fatal failure} is any failure that renders a system unable to perform its duties without external intervention, and is most commonly caused by environmental perturbations such as noise, dim or bright lighting, blurred frames, as well as short or long-term scene changes (\textit{e.g.} dynamic objects). 
While some use cases only require episodic or short-term operation, many applications call for long-term deployment: home maintenance, autonomous inspection of industrial facilities, and so on. In the context of robot navigation, we refer to such long-term operation as \textit{Lifelong SLAM}. Given the current capabilities and performance described above, we believe that the success of Lifelong SLAM is primarily dependent on the capacity of a system to be generally robust with respect to perturbations which may not be known a priori.

Previous efforts in evaluating the robustness of multiple SLAM systems have focused on specific types of perturbations \cite{park2017icra, rosler2020dynamic}, often limited to a specific sensing modality \cite{lomps2020evaluation, prokhorov2019measuring}, without considering whether building in resilience against specific perturbations may incur any trade-offs with respect to other challenging factors or measuring the general robustness of the system. 
Our work addresses this gap by introducing an evaluation methodology for assessing the robustness of SLAM solutions supporting various sensing modalities and degrees of freedom, in the presence of a variety of perturbations evaluated independently as well as in combination. We demonstrate the validity of our approach by performing an extensive evaluation of 6 SLAM systems (Table \ref{tab:slam_algorithms}) on 6 datasets (Table \ref{tab:slambench_datasets}) across 3 computing platforms, in both episodic and long-term operation settings. 
Figure \ref{fig:samples} contains a selection of frames with occlusions and dynamically-moving elements, illumination changes in real and synthetic scenes, frames from drone sequences containing motion blur and no reliable features, lifelong operation challenges, colour frames containing lighting differences, blur, and dynamic objects. The accompanying video shows a qualitative comparison of 4 algorithms running on a sequence with dynamic elements.

\section{Related Work}

While the problem of robustness has been acknowledged since the early days of SLAM \cite{folkesson2004robust, levinson2010robust, Cadena16tro-SLAMfuture}, it remains one of the most significant challenges. We briefly review the literature on SLAM robustness with respect to perturbations relevant to our work.

\noindent \textbf{Illumination changes} may occur due to natural (\textit{e.g.} varying sunlight) or artificial causes (\textit{e.g.} blinking lightbulbs), translating into sudden changes in image brightness, either locally or globally. A large number of works rely on brightness constancy for mapping \cite{whelan2015elasticfusion, kerl2013dense, forster2014svo, engel2014lsd}, and may be negatively affected by such changes. Methods to improve robustness to illumination changes include active exposure control
\cite{zhang2017active,shim2018gradient,kim2017robust}, binary local descriptors for brightness normalization such as \textit{Census transform} \cite{alismail2016direct,alismail2016lowlight}, while other works developed illumination-invariant metrics to register images \cite{pascoe2017nid, wang2007non}. A detailed evaluation of the performance of direct methods under such perturbations is presented in \cite{park2017icra}, whose dataset we adopt. 

\noindent \textbf{Dynamic elements} are one of the most widely encountered type of perturbation: virtually all settings where SLAM is employed, from home robots to autonomous vehicles to augmented reality are bound to feature movement. Over the years, a number of solutions have been proposed \cite{saputra2018visual}. Given that the static part of a scene provides the most reliable information for computing the camera pose, many approaches to dynamic SLAM attempt to segment the input into static and dynamic parts. Methods include the use of optical flow \cite{derome2015moving,scona2018staticfusion,cheng2019improving}, geometric constraints \cite{tan2013robust}, alignment residuals \cite{palazzolo2019refusion}, and semantic information \cite{bujanca2019fullfusion,bescos2018dynaslam,yu2018ds,sheng2020dynamic,xiao2019dynamic,mu2019visual}. 

\noindent \textbf{Fast camera movement} on robots and drones often results in motion blur, hindering both feature detection and direct alignment, methods widely employed by SLAM systems. \cite{pretto2009visual,lee2011simultaneous,mustaniemi2018fast} use frame deblurring to ensure reliable features can be identified; FLaME \cite{greene2017flame} proposes to use low quality but high frequency depth estimation to aid obstacle avoidance in drone flight.

\noindent \textbf{Lifelong SLAM} and long-term localisation are long-standing problems \cite{tipaldi2013lifelong, johannsson2013toward, kretzschmar2010lifelong, einhorn2015generic}. In the past year, new benchmarks challenging the state-of-the-art have appeared \cite{longtermlocalizationchallenge, shi2020we}, and promising results (usually based on detecting learned features) have been proposed for localisation \cite{sarlin2020superglue, Torii-CVPR2015, taira2018inloc} as well as SLAM \cite{li2020dxslam}. We reuse the dataset and metrics of the Lifelong Robotic Vision challenge in our evaluation \cite{shi2020we}.

We aim for a comprehensive evaluation of the robustness of SLAM systems, but recognise that other factors, such as weather \cite{porav2019can, porav2019don} or limited visibility \cite{kim2013real} are also of practical importance. Our methodology should help evaluate such perturbations in the future, as well as other factors (e.g.\ 3D reconstruction, semantic labelling).

\section{Methodology}
\subsection{Evaluation workflow}

We design our pipeline to support single and multi-sequence inputs and use the latter for Lifelong SLAM evaluation. Our evaluation pipeline adopts and extends the tools for trajectory alignment, visualisation, and metric computation provided by the open-source SLAMBench framework\footnote{Code available at \href{https://github.com/pamela-project/slambench}{https://github.com/pamela-project/slambench}}\cite{bodin2018slambench2,bujanca2019slambench}. 
The software containing routines for configuring and initialising each system, streaming data into the algorithm and collecting outputs (estimated pose and monitoring the state of the system) will be made public. Importantly, preparing an algorithm for evaluation only involves writing a thin wrapper around each algorithm and does not require modifying the code of the system.

To assess the accuracy of each algorithm, we use the Absolute Trajectory Error (ATE) and Relative Pose Error (RPE) introduced in the TUM RGB-D \cite{sturm12iros} dataset. In contrast to most evaluation procedures where the alignment and computation of the metrics is done only using the final trajectory, we continuously monitor the ATE and RPE by realigning the trajectories in $SE(3)$ using Umeyama's method \cite{umeyama1991least} and measuring the errors every time the SLAM system outputs a new pose. To prevent algorithms from dropping frames, new data is sent \textit{after} the previous frame finished processing.

Figure \ref{fig:correlate} shows spikes in error correlated with the input data causing them, allowing us to deduce the particular sensitivities of individual algorithms, as well as to identify scenes which are generally challenging for SLAM algorithms.
Since performing these routines for every frame can be expensive and could affect measurements, we use existing mechanisms in SLAMBench to report execution times and resource usage by the SLAM algorithms independently of evaluation and trajectory alignment.

\subsection{Lifelong SLAM}

Evaluating Lifelong SLAM entails simulating common long-term operation scenarios. Each algorithm is fed multiple sequences captured in the same environment, with aspects such as initial position, time of day, lighting, and so on, varying across sequences.

In addition to computing the per-sequence ATE and RPE, the metric Correct Rate of Tracking (CRT) is adopted \cite{shi2020we}.
Environmental perturbations may cause SLAM algorithms to lose tracking. The ATE may be unevenly affected by a loss of tracking; e.g.\ losing tracking in the late stages of a sequence could have a significant impact on the Mean ATE. On the other hand, the CRT metric measures the ratio of correct tracking time with respect to the whole time span of the data. Correctness of each estimated pose can be determined with user-specified thresholds of ATE and other per-frame metrics. By combining the two metrics, we can better capture the overall performance, tracking failures, as well as the time spent correctly tracking the camera pose.

\section{Experiments} \label{sec:experiments}

\subsection{Experimental setup}
We perform the experiments on the following three hardware platforms (\textit{Workstation}, \textit{Laptop}, and \textit{Jetson}), running under 64-bit Ubuntu 18.04 OS:

The workstation is a desktop with 32 GB of RAM, a 14-core Intel Core i9-9940X chip (3.30GHz), and an Nvidia TITAN RTX GPU with 24GB VRAM and 4608 CUDA cores. The laptop is a Lenovo ThinkPad P53 with 16GB of RAM, a 6-core Intel Core i7-9850H (2.60GHz), and an Nvidia Quadro RTX 3000 with 1920 CUDA cores and 6GB of VRAM. The Jetson is an 
Nvidia Jetson Xavier AGX. This is a platform commonly used in ground robots, featuring a 8-core ARMv8.2 64-bit CPU (2.25GHz), 16 GB of RAM, and a 512-core Nvidia Volta GPU. 
The device is set up to deliver the maximum performance, with a peak power use of 30W. 

To control for any differences not inherent to the algorithms, we ensure that, wherever possible, on each platform any common dependencies undertaking significant computational tasks, such as \textit{OpenCV} or \textit{g2o}, are fixed to the same version across all the SLAM systems evaluated. We use \textit{gcc 7} for compilation across all algorithms and platforms, and CUDA 10.2 for GPU-based implementations. The DVFS of the processing cores (Turboboost) and GPU (adaptive clocking) are disabled. 
All the build processes have been modified to use the highest levels of compiler optimisation. The hyperparameters of SLAM systems are configured following the recommendations of the original papers/repositories, if available, or otherwise using the default settings.

Using the appropriate input modalities provided by each dataset (Table \ref{tab:slambench_datasets}), we evaluate 6 open-source SLAM systems selected to cover a diversity of designs with respect to input modalities and map representations.

\textbf{OpenVINS} \cite{Geneva2020ICRA} is a stereo visual-inertial SLAM system which uses an Extended Kalman Filter to fuse visual odometry with inertial measurements.

\textbf{ORB-SLAM2} \cite{murORB2} is a popular real-time SLAM system based on sparse ORB features. It incorporates RGB-D, monocular and stereoscopic input modalities.

\textbf{ORB-SLAM3} \cite{ORBSLAM3_2020} is a recently released SLAM system developed on top of ORB-SLAM2 which introduces a multiple map system and visual-inertial odometry to improve robustness.

\textbf{ElasticFusion} \cite{whelan2015elasticfusion} provides a globally-consistent dense RGB-D reconstruction approach that does not require a pose graph and represents the map using fused surfels \cite{pfister2000surfels}.

\textbf{FullFusion} \cite{bujanca2019fullfusion} is a framework for semantic reconstruction of dynamic scenes. FullFusion leverages semantic information to separate RGB-D inputs into a static and a dynamic frame. A modified implementation of KinectFusion is used to compute the pose and reconstruct a semantically labelled model of the static scene elements. 

\textbf{ReFusion} \cite{palazzolo2019refusion} is a dense RGB-D 3D reconstruction method which exploits residuals obtained after the registration of input data with the reconstructed model to identify and filter out dynamic elements in the scene.

The datasets have been chosen to cover a wide range of conditions common in SLAM applications:
\begin{itemize}
    \item Camera motion / hardware platform: ground robot, aerial vehicle, handheld sensor, linear motion (in synthetic scenes).
    \item Scene type: synthetic, indoor, outdoor, empty corridors, busy market or cafe.
    \item Lighting: sudden exposure changes, daylight, night, continuously changing local and global illumination, flashlights.
    \item Movement: varying levels of movement, both rigid and non-rigid. All combinations of static and moving camera with static and moving scenes.
    \item Sensors: RGB-D, Stereo cameras, IMU, Wheel odometry, Sequences featuring sensor degradation.    
\end{itemize}

An experiment refers to the execution combining one SLAM system (Table \ref{tab:slam_algorithms}) and one sequence of a given dataset (Table \ref{tab:slambench_datasets}).
Each experiment is performed 10 times on each of the 3 platforms generating a total of approximately 20000 data points. 

Given the large number of experiments and the necessity to differentiate by platform and perturbation, the paper contains only a subset of the results and metrics. Full data is available on the website\footnote{\href{https://robustslam.github.io/evaluation}{https://robustslam.github.io/evaluation}}. We adopt the following strategy to present aggregated data under each setting: for each sequence, we compute the median of the translational ATE-RMSE over the 10 runs, normalised by the metric length of the sequence to ensure equal weighting across sequences.
Note that the aggregate plots may not always be representative of the performance on individual sequences. 

\begin{figure}[tb!]
   \centering
      \includegraphics[width=\columnwidth]{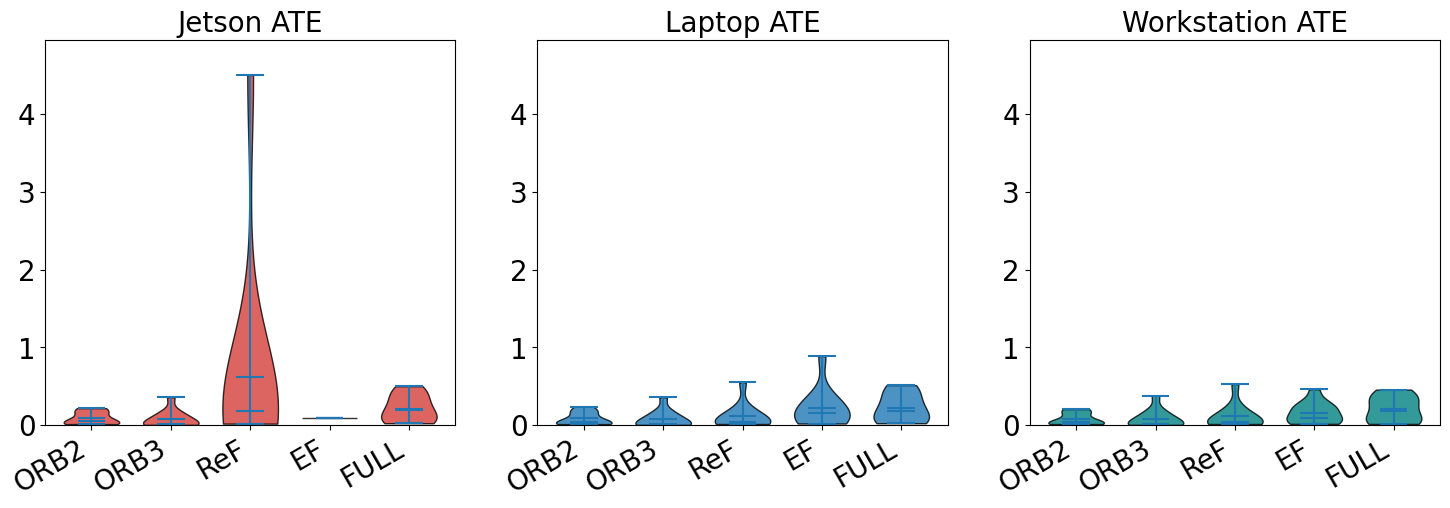}
      \scriptsize{(a) TUM}
      \includegraphics[width=\columnwidth]{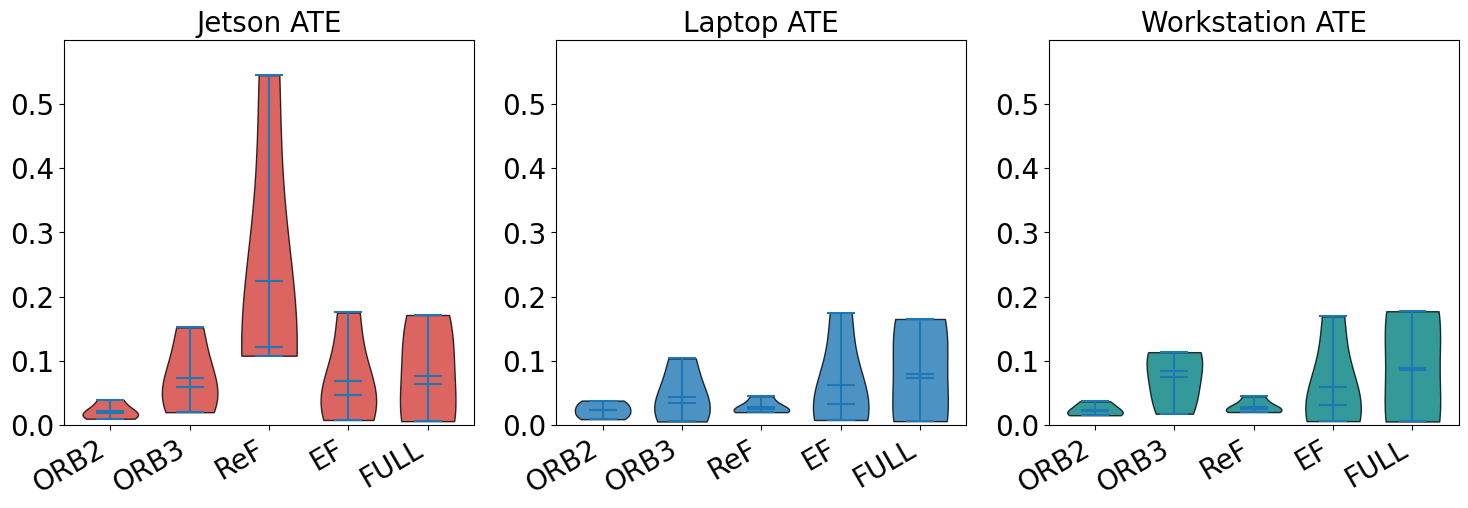}
      \scriptsize{(b) ICL-NUIM}
      \includegraphics[width=\columnwidth]{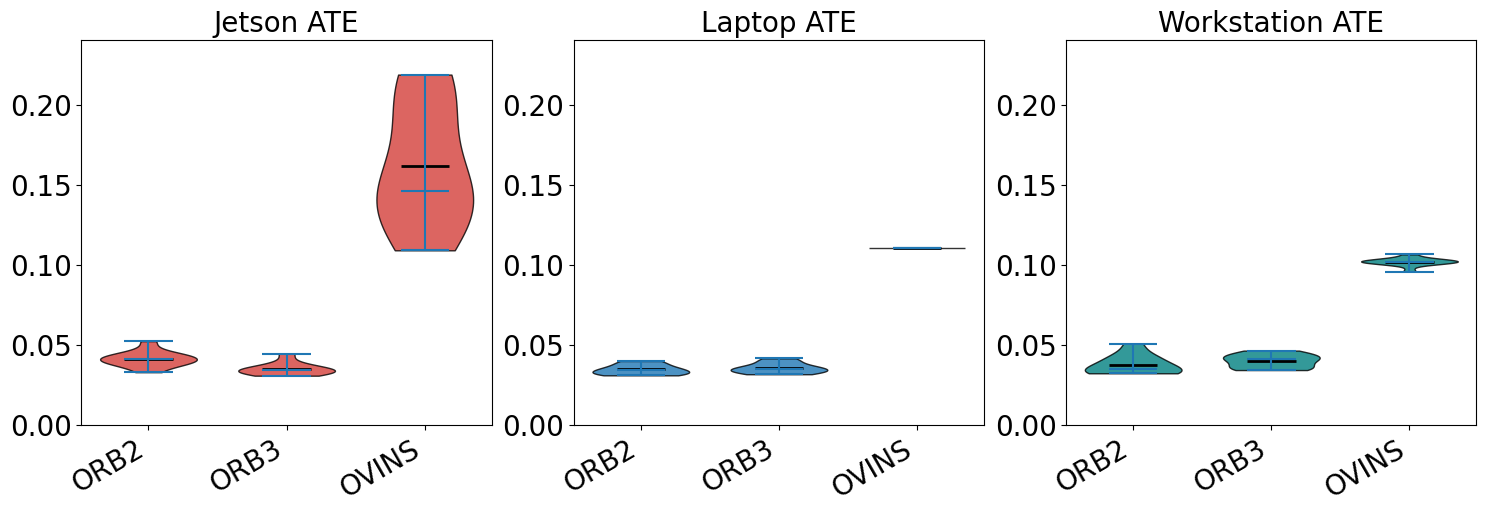}
      \scriptsize{(c) EuRoC-MAV}
      \caption{Baseline performance results.}
   \label{fig:tum}
    \vspace{-2em}
\end{figure}

\subsection{Results}

\noindent \textbf{Baseline performance} -- We evaluate the trajectory estimation accuracy of each SLAM system on selected sequences of widely-adopted datasets where no significant perturbations are present. The RGB-D based SLAM systems are evaluated with 12 sequences from the TUM \textit{freiburg1} and \textit{freiburg2} datasets \cite{sturm12iros} and the 4 sequences of the ICL-NUIM \textit{living room} dataset \cite{iclnuim}. Our results (Figures \ref{fig:tum}-a and \ref{fig:tum}-b) are consistent with the existing literature. ORB-SLAM2, ORB-SLAM3 and ElasticFusion are accurate within 1\% on all sequences and no individual runs exceeded 3\% error. FullFusion and ReFusion maintained their ATE below 3\% on most runs, but scored worse than the aforementioned systems (with few exceptions). ORB-SLAM3 is the most accurate in this baseline setting, with ORB-SLAM2 closely after. 

SLAM systems supporting stereo and visual-inertial SLAM are evaluated on the 7 \textit{easy} and \textit{medium} sequences of the EuRoC-MAV dataset. Figure \ref{fig:tum}-c shows all 3 SLAM systems have similar accuracies and performed within a 0.5\% error margin. Unexpectedly, on some of the \textit{machine hall} sequences, ORB-SLAM3, using the stereo VIO mode, performed slightly worse than ORB-SLAM2 in stereo mode.

\begin{figure}[tb!]
  \centering
      \includegraphics[width=\columnwidth]{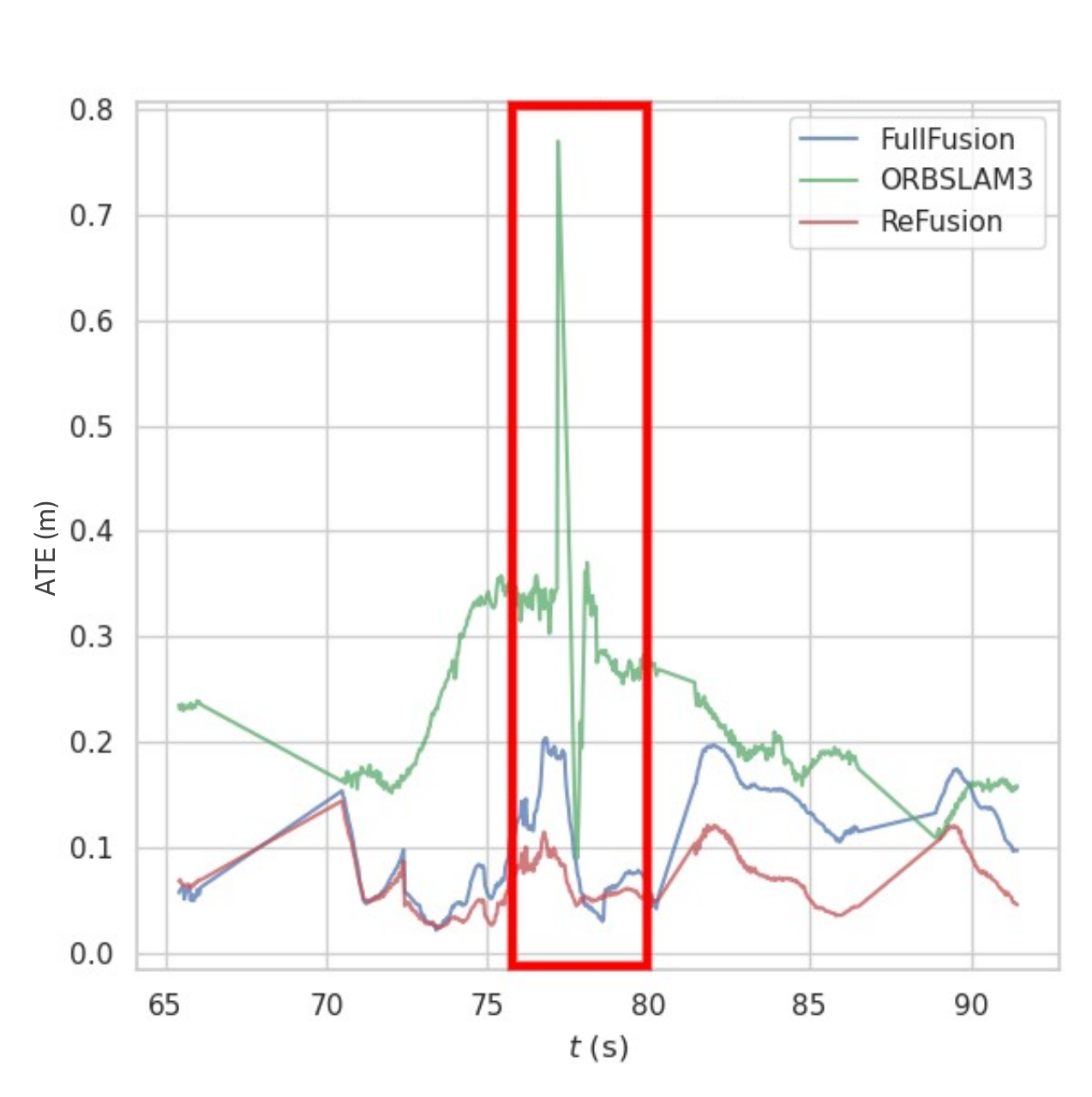}
      \caption{ReFusion, ORB-SLAM3 and FullFusion executing the \textit{moving\_nonobstructing\_box} sequence of the Bonn dataset. The red rectangle highlights the period of time when a person enters the scene, moves a box and leaves.
      }
  \label{fig:bonnspikey}
    \vspace{-1em}
\end{figure}
\begin{figure*}[tb!]
  \centering
  \includegraphics[width=0.9\textwidth]{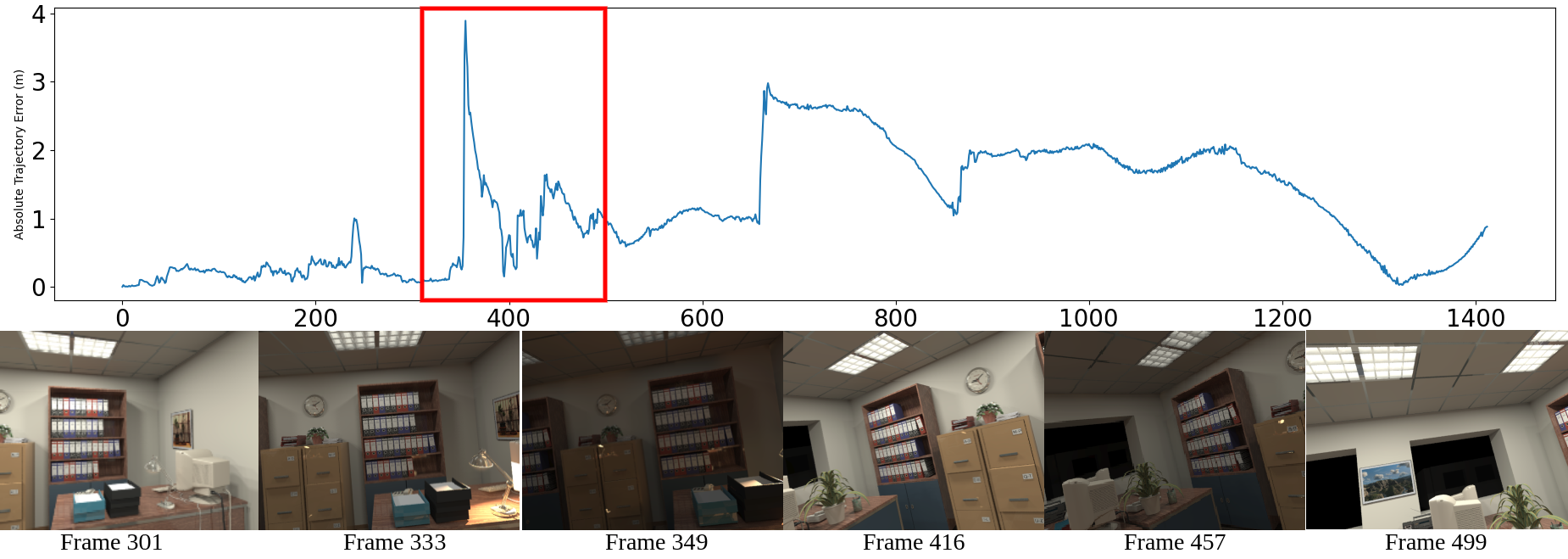}
  \caption{Single run of ReFusion on the \textit{syn2} sequence with both local and global illumination changes. Significant spikes in error occur during changes in illumination (top). Bottom: frames corresponding to the highlighted area.}
    \label{fig:correlate}
    \vspace{-1em}
\end{figure*}

\noindent \textbf{Illumination changes} -- We use the ETH Illumination dataset to analyse the resilience of SLAM systems using 3 real and 10 synthetic RGB-D sequences. The dataset features multiple types of illumination change: local, global, local and global, and flashlight. 
The real sequences are captured with handheld Kinect v1 sensor, in an environment closely resembling the TUM RGB-D setting. The synthetic scenes are adapted from the ICL-NUIM dataset. 
Thanks to the illumination invariance of ORB features, both ORB-SLAM2 and ORB-SLAM3 appear to be unaffected by any type of illumination change, obtaining similar scores to the baseline TUM and ICL-NUIM. In contrast, ElasticFusion and ReFusion use photometric errors which assume constant illumination, leading to high error rates. Figure \ref{fig:correlate} highlights the effects of changes in illumination on ReFusion.

\noindent \textbf{Dynamic elements} --- We use the 24 dynamic sequences in the Bonn RGB-D Dataset.
All scenes were captured in the same space and include people handling objects such as boxes and balloons. Varied levels of movement are present, ranging from mostly static scenes to complete occlusion of the background by moving objects for extended periods. 

Having been published together with the Bonn dataset, ReFusion performs best in the presence of dynamic elements. 
Nonetheless, ORB-SLAM2 and ORB-SLAM3 perform more accurately than ReFusion on scenes with negligible movement or when the dynamic elements are untextured, relying mostly on background keypoints, and are able to recover when dynamic objects briefly enter and leave the frame, but fail under severe motion.
Figure \ref{fig:bonnspikey} highlights the moderate increase in error in ReFusion and FullFusion compared to a significant spike in error for ORB-SLAM3. FullFusion's segmentation module relies on semantics to remove dynamic objects from frames. As such, FullFusion performs well only when recognized classes are present in the scene (\textit{person}), but is highly sensitive to any other movement, often experiencing failures (\textit{balloon, box} sequences), unlike more general algorithms such as ReFusion.
As expected, due to using all the data in the frame and assuming only camera movement, ElasticFusion is severely affected, with a noticeable drop in accuracy occurring as soon as a dynamic object enters the scene, without subsequent recovery, and fails entirely on the highly dynamic scenes.

\noindent \textbf{Lifelong SLAM} --- 
OpenLORIS-Scene is a comprehensive dataset featuring a total of 22 sequences captured in 5 common environments (office, corridor, home, cafe, and market) at different times of the day using commercial service robots. Compared to most SLAM datasets which often present tightly controlled scenarios, OpenLORIS contains realistic settings for service robots, and a wide variety of challenging factors: occlusions, dynamic motion, featureless areas, and lighting changes. 

\begin{figure*}[tb!]
  \centering
      \begin{tabular}{cccccc}
        \includegraphics[width=2.5cm]{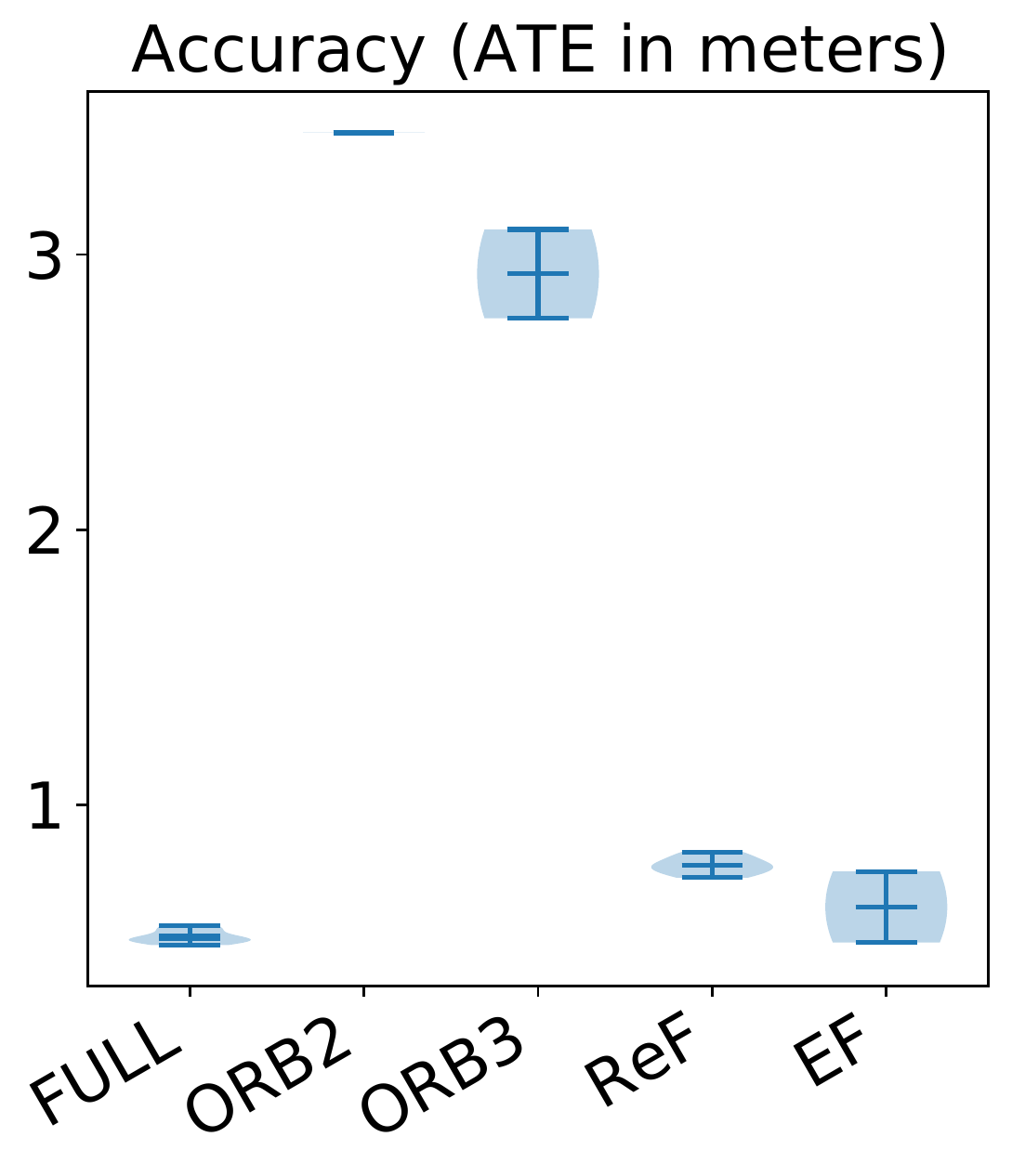} &
        \includegraphics[width=2.5cm]{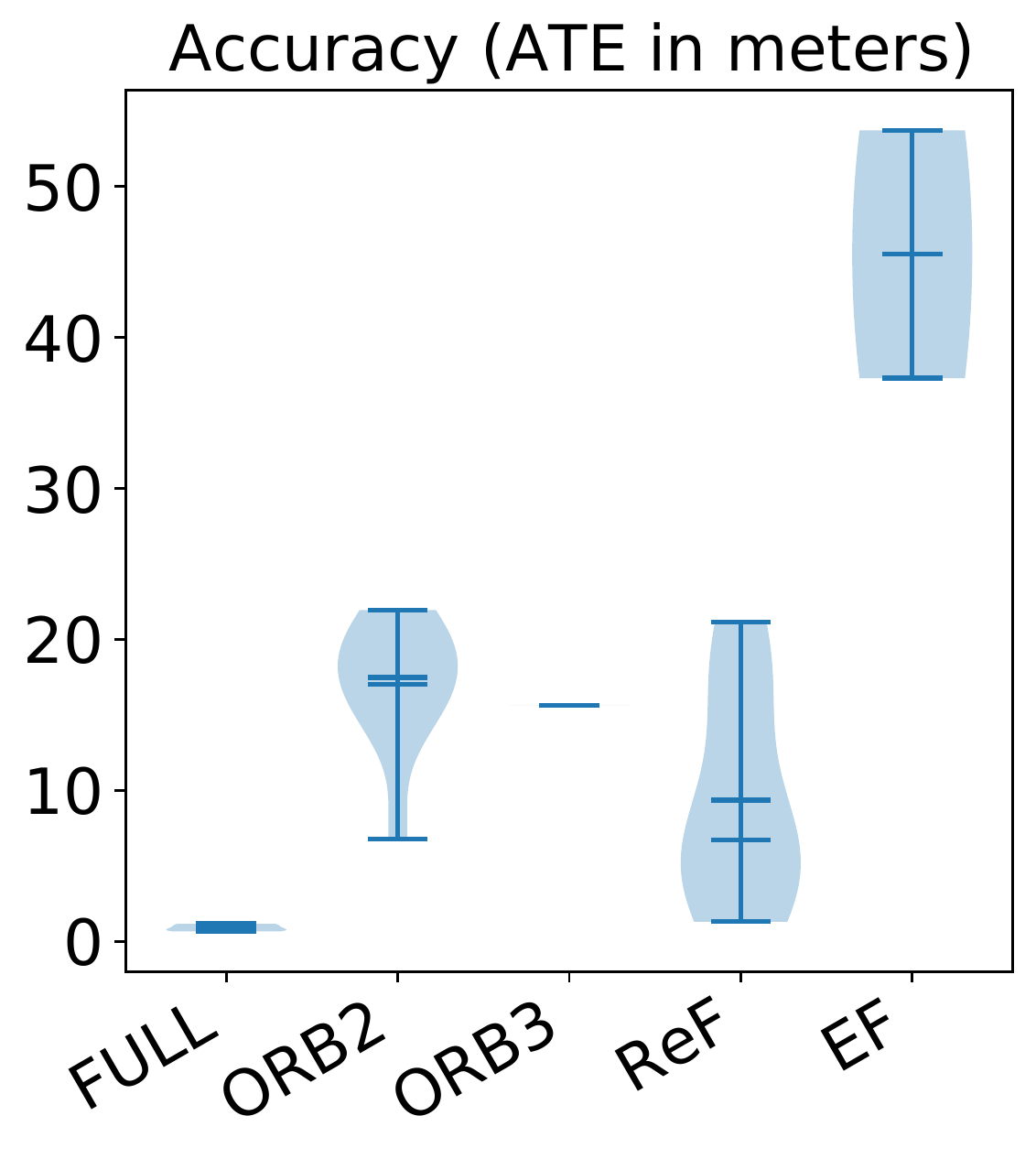} & 
        \includegraphics[width=2.5cm]{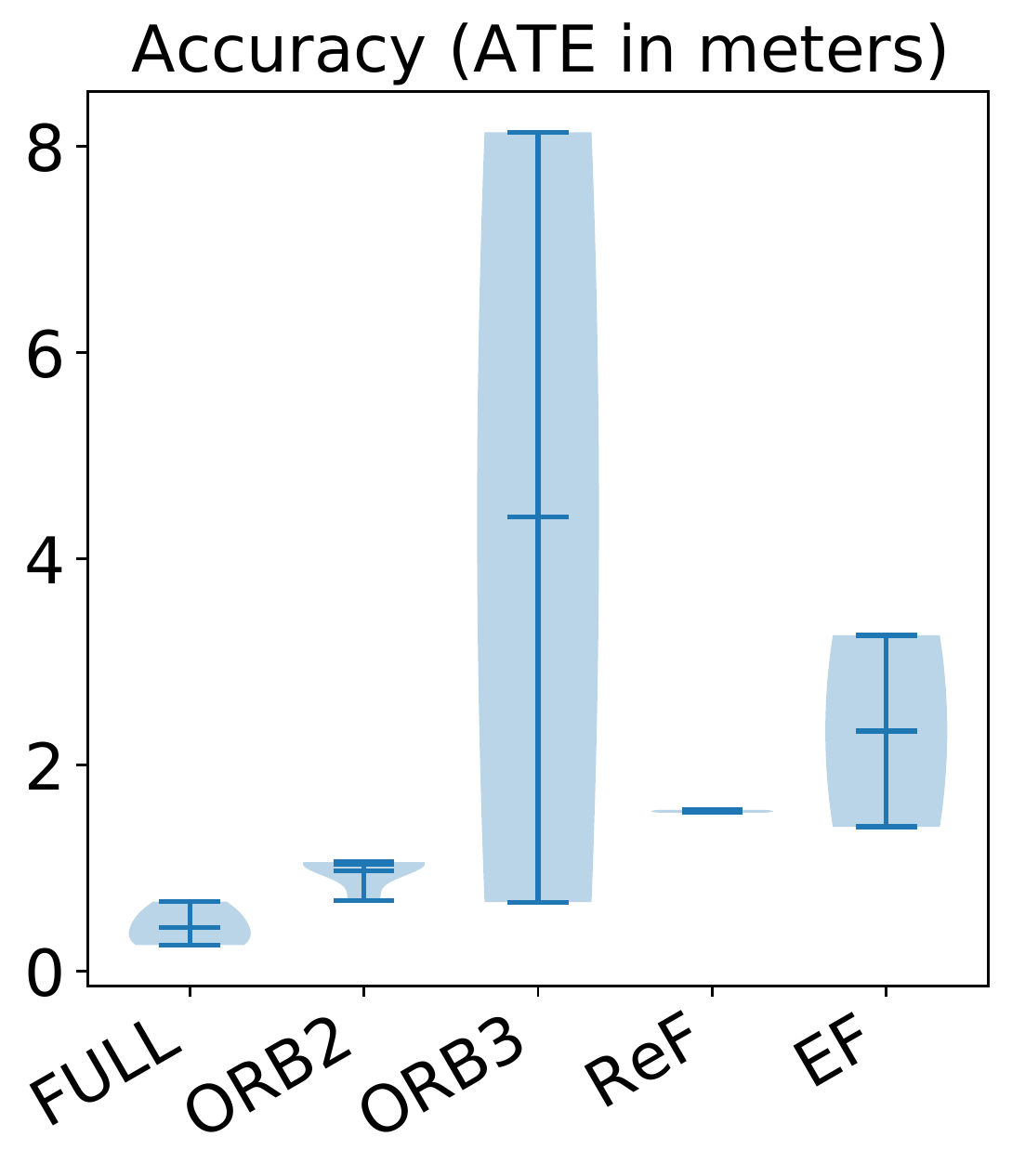} &
        \includegraphics[width=2.5cm]{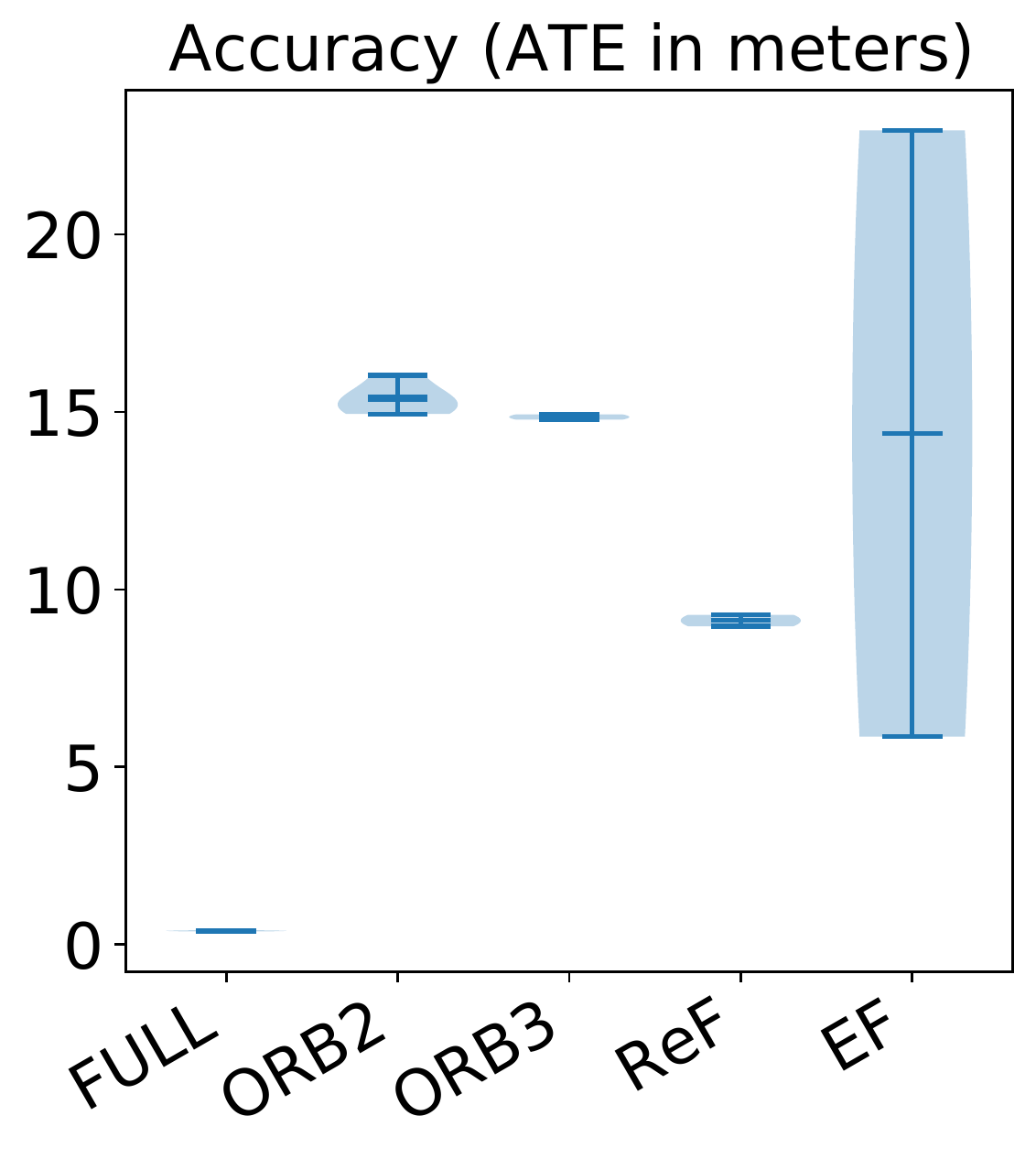} &
        \includegraphics[width=2.5cm]{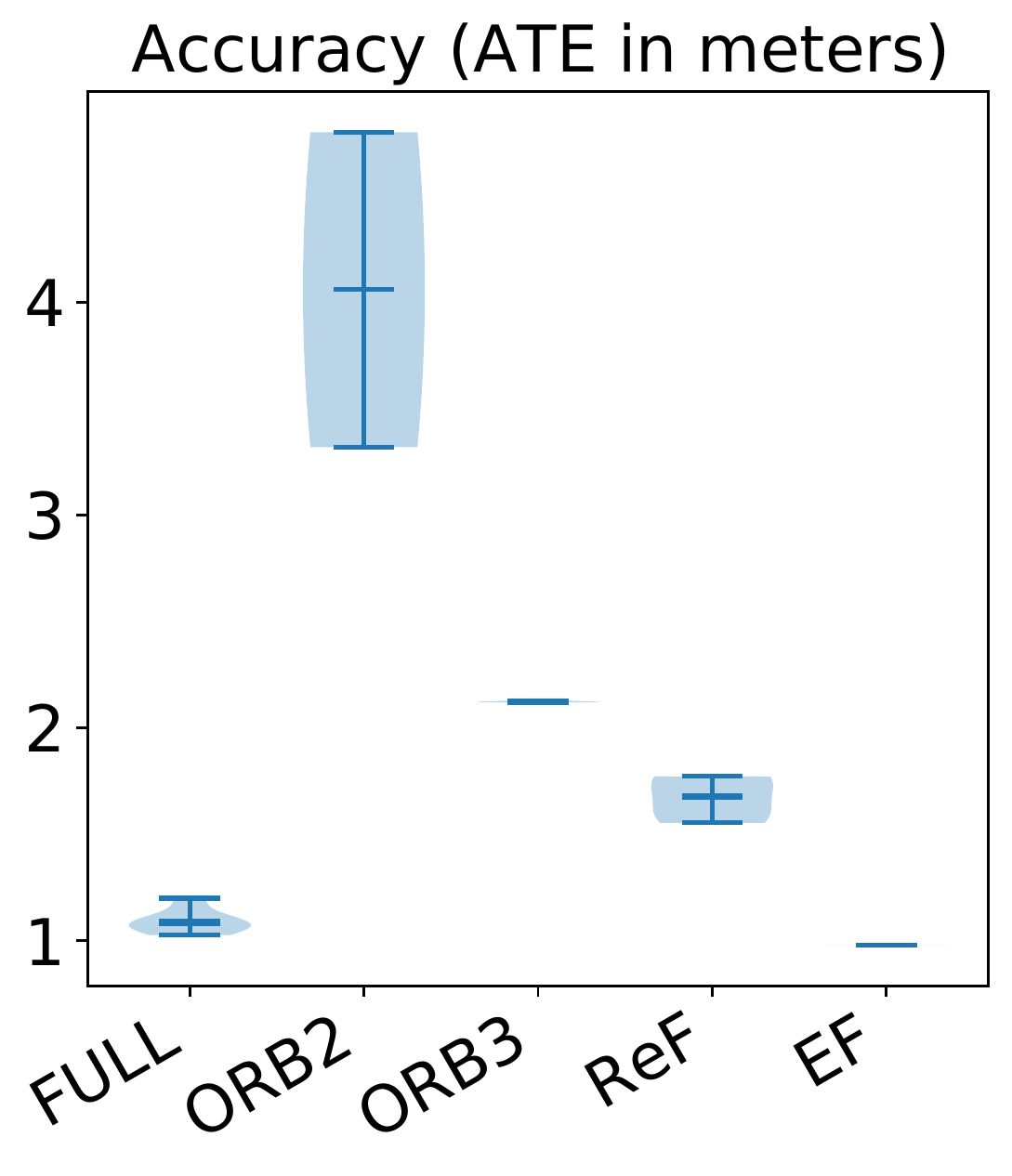} &
        \includegraphics[width=2.5cm]{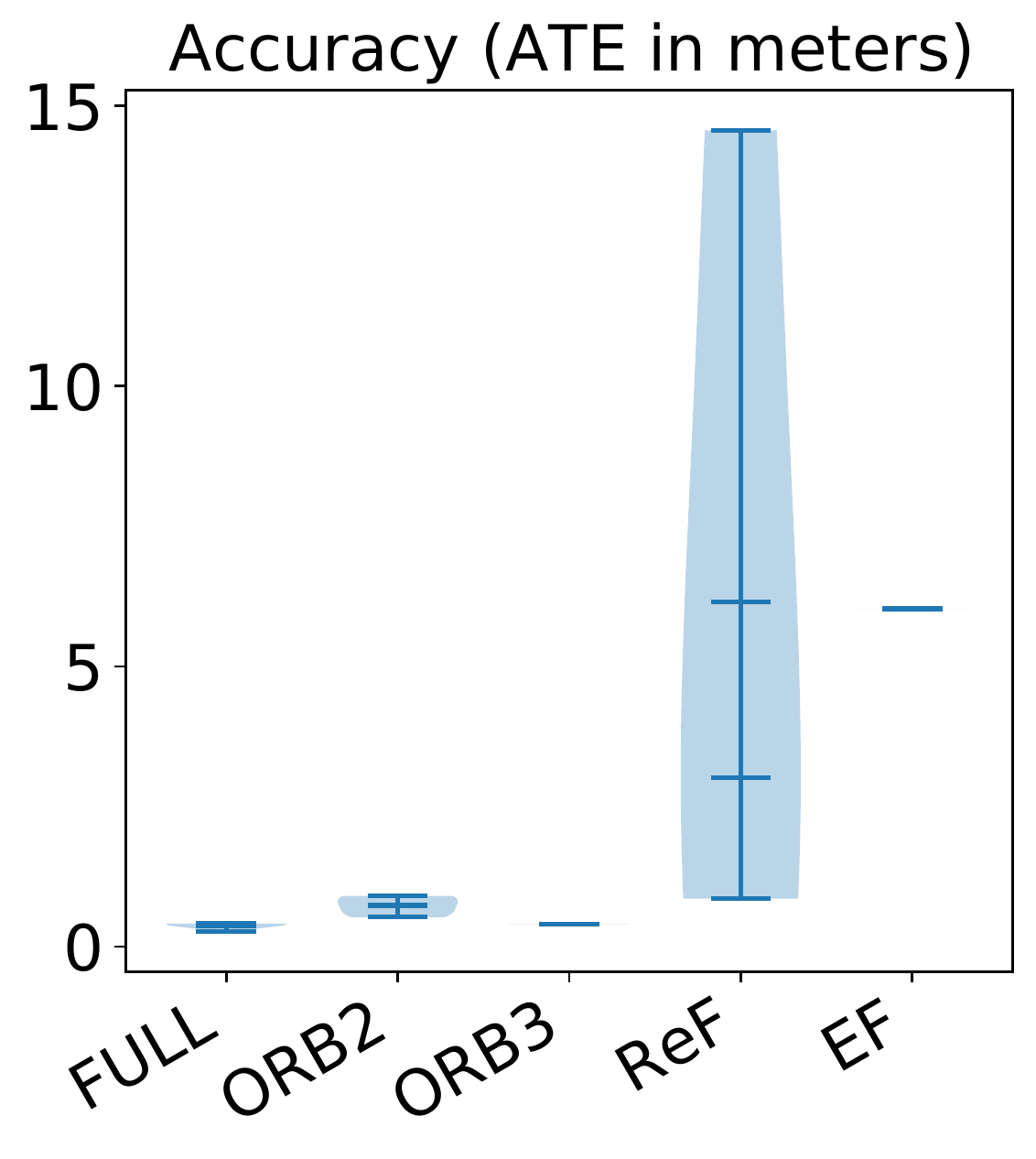} \\
        \footnotesize{cafe2} & \footnotesize{corridor1} & \footnotesize{corridor4} & \footnotesize{corridor5} & \footnotesize{home2} & \footnotesize{market2}\\

        \includegraphics[width=2.5cm]{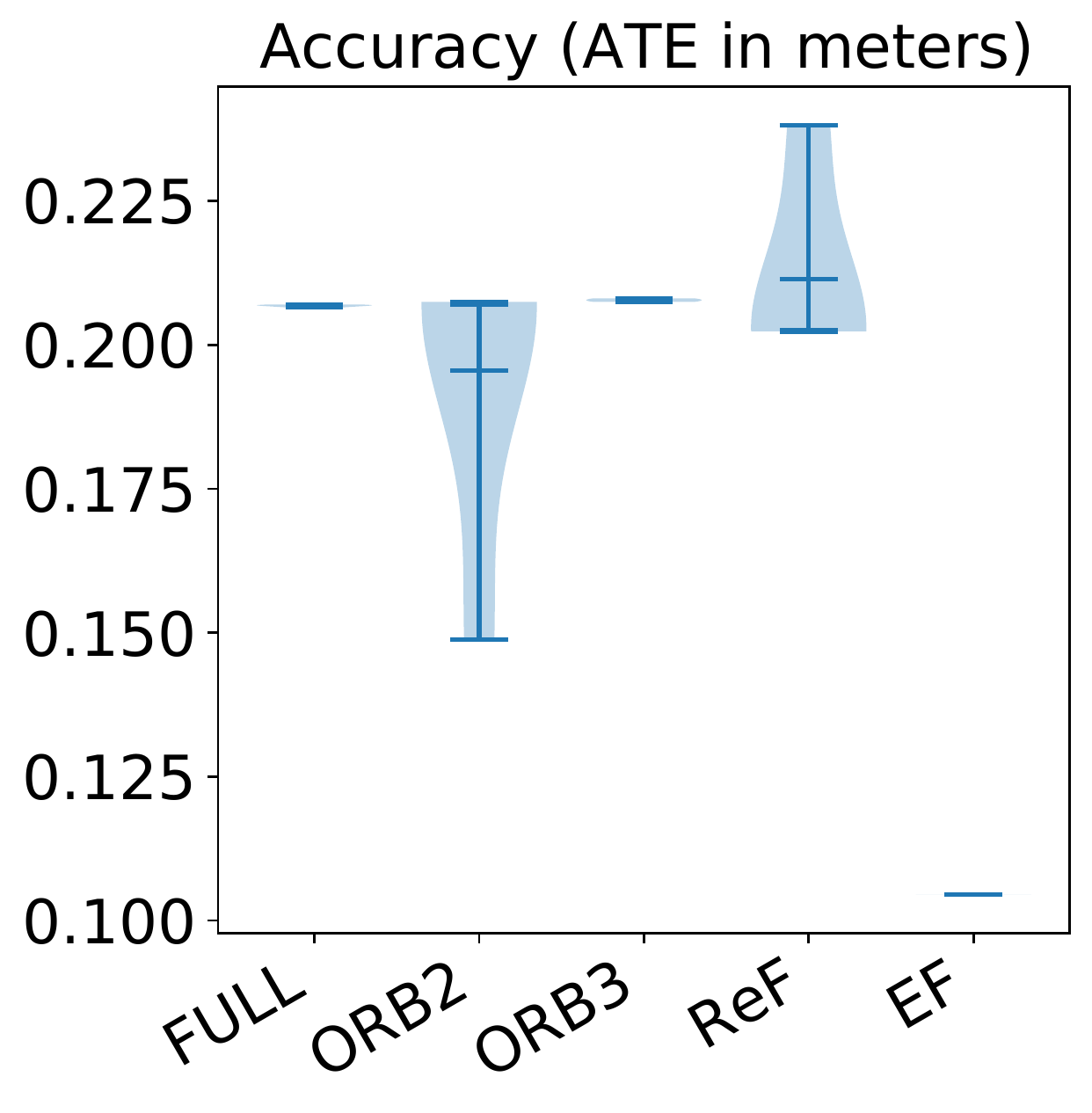} &
        \includegraphics[width=2.5cm]{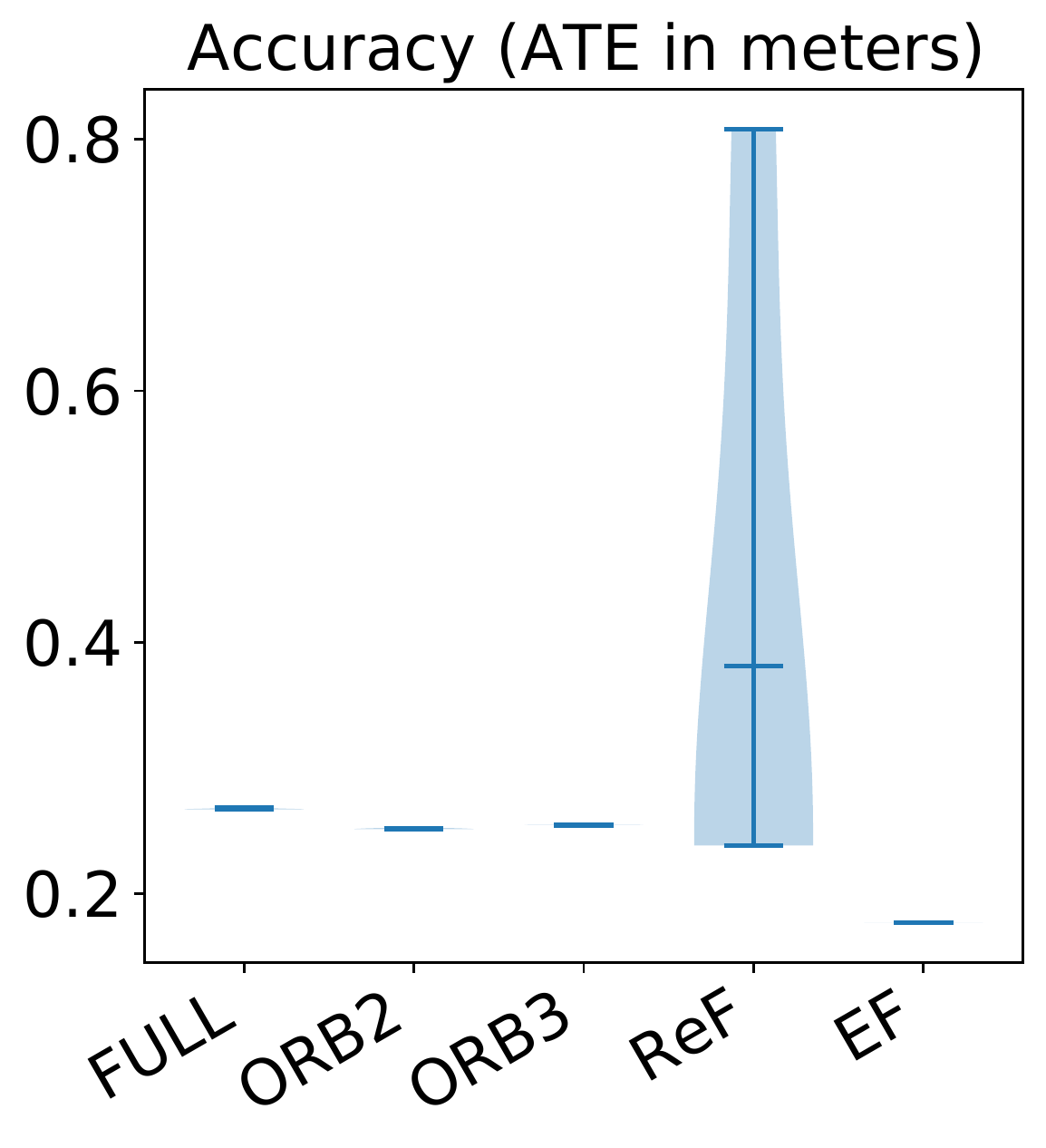} &
        \includegraphics[width=2.5cm]{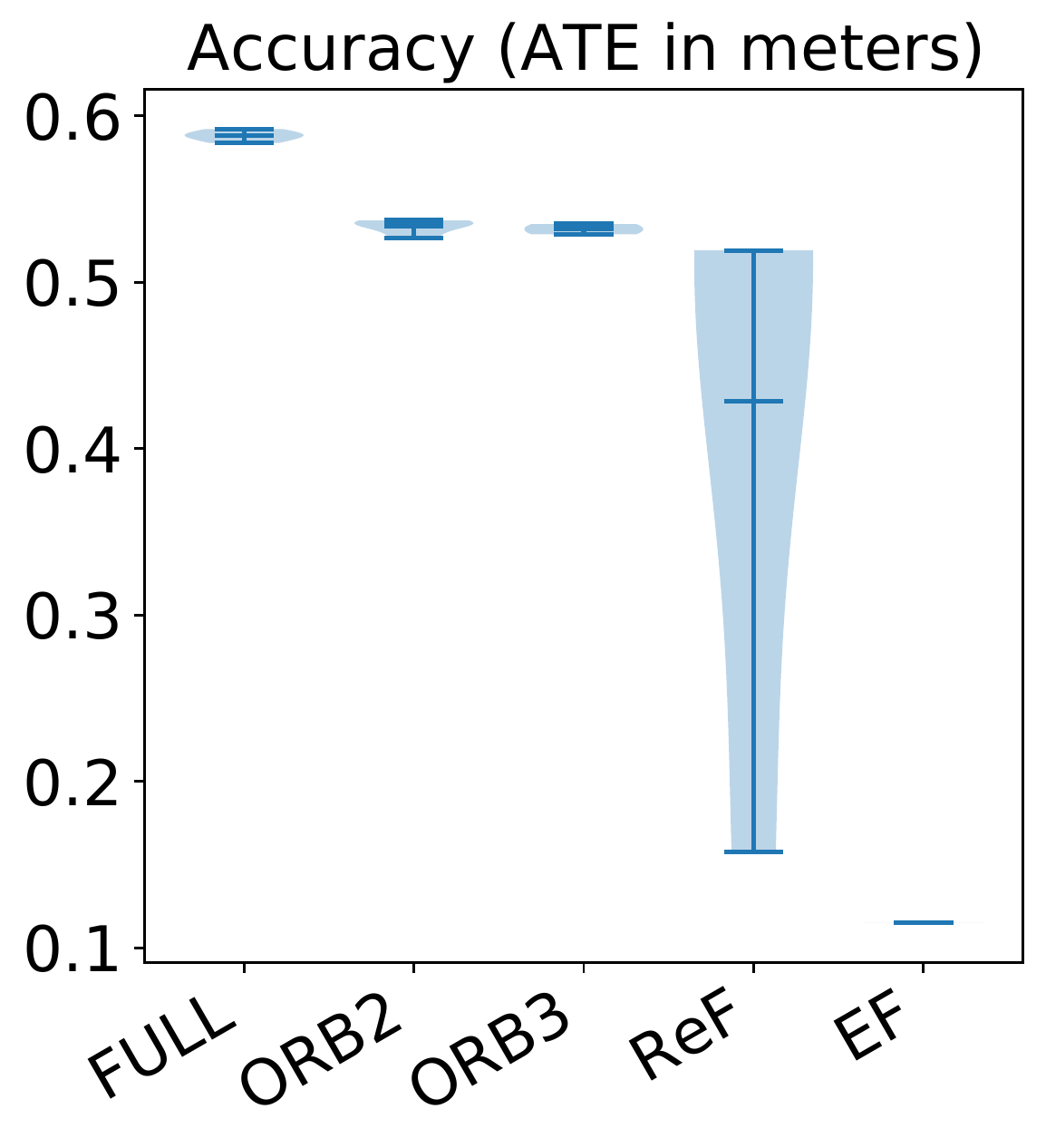} &
        \includegraphics[width=2.5cm]{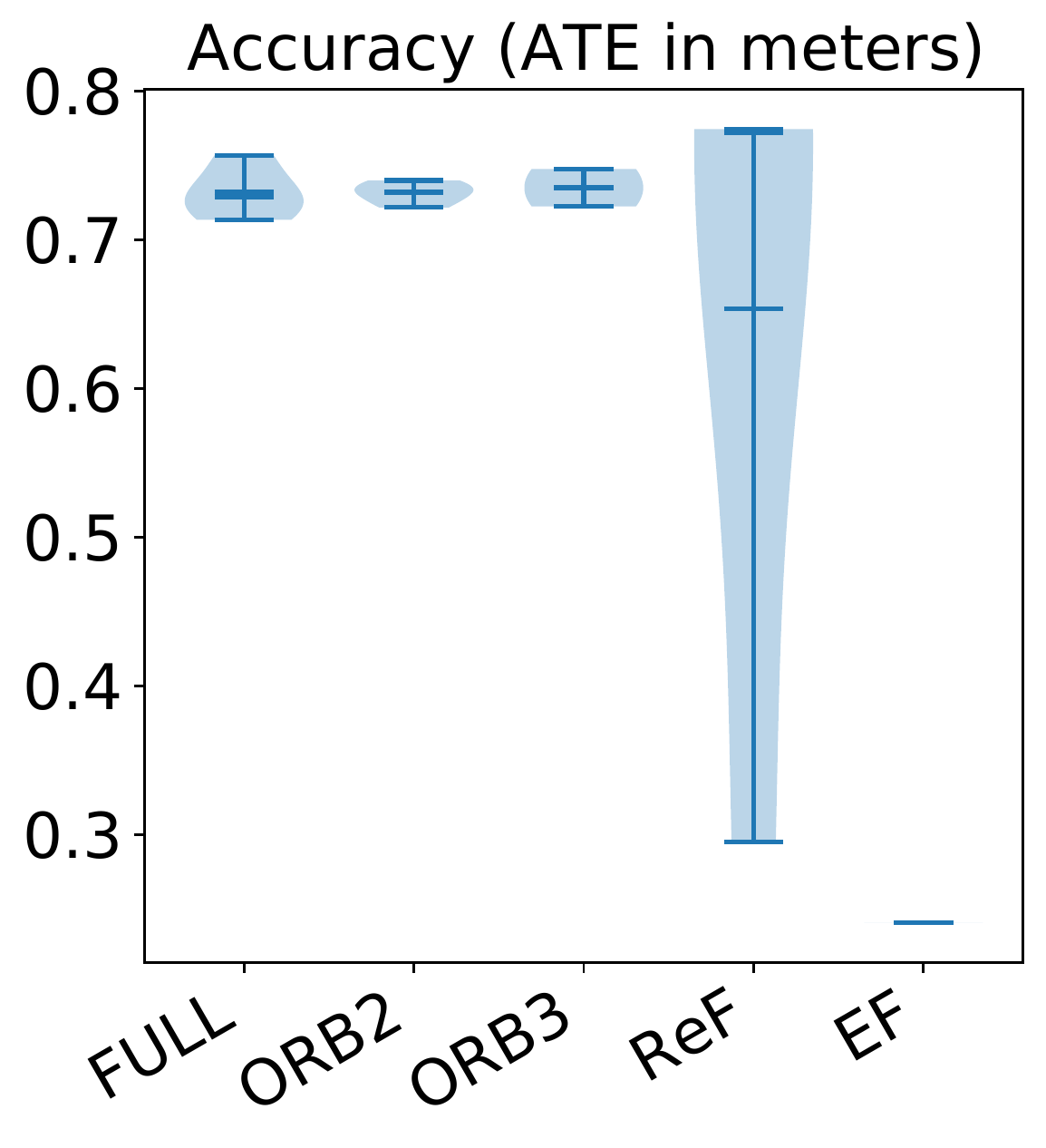} &
        \includegraphics[width=2.5cm]{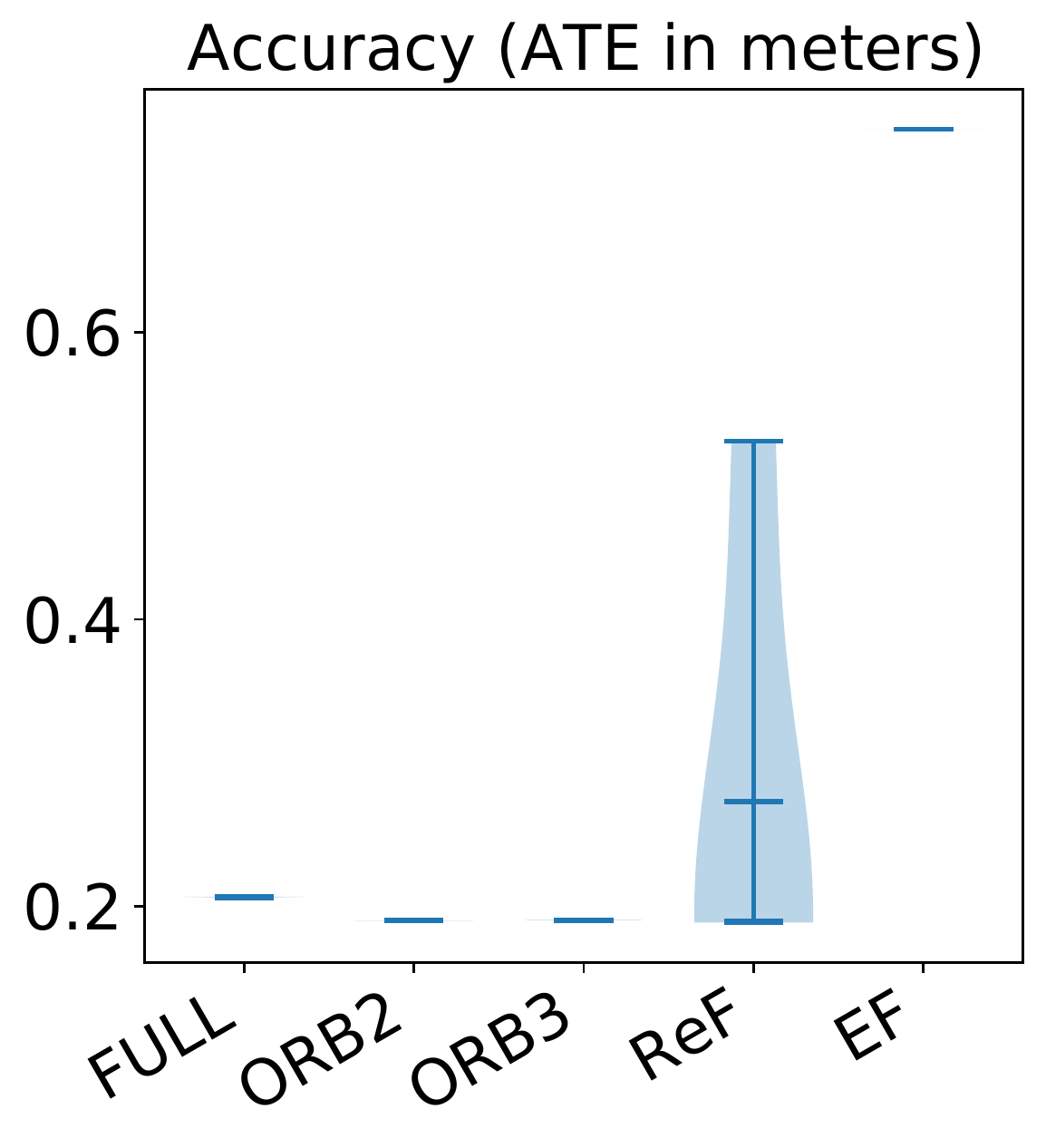} &
        \includegraphics[width=2.5cm]{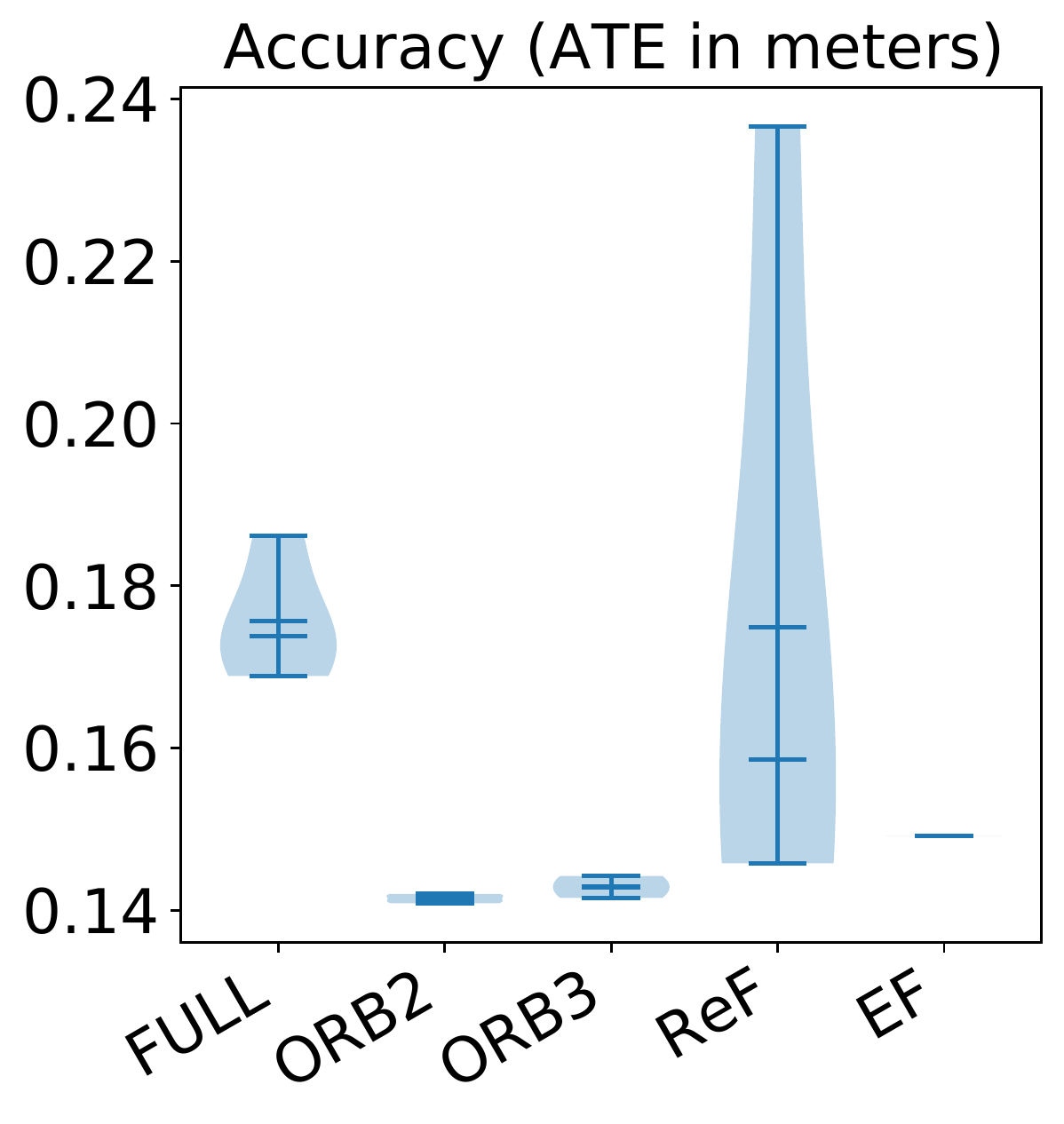}
         \\ 
        \footnotesize{office1} & \footnotesize{office2} & \footnotesize{office4} & \footnotesize{office5} & \footnotesize{office6} & \footnotesize{office7} \\ 
        
      \end{tabular}
      \caption{OpenLORIS -- accuracy results for a subset of scenes.}
  \label{fig:openloris_ATE}
  \vspace{-1em}
\end{figure*}

\begin{figure*}[tb!]
  \centering
      \includegraphics[width=\textwidth]{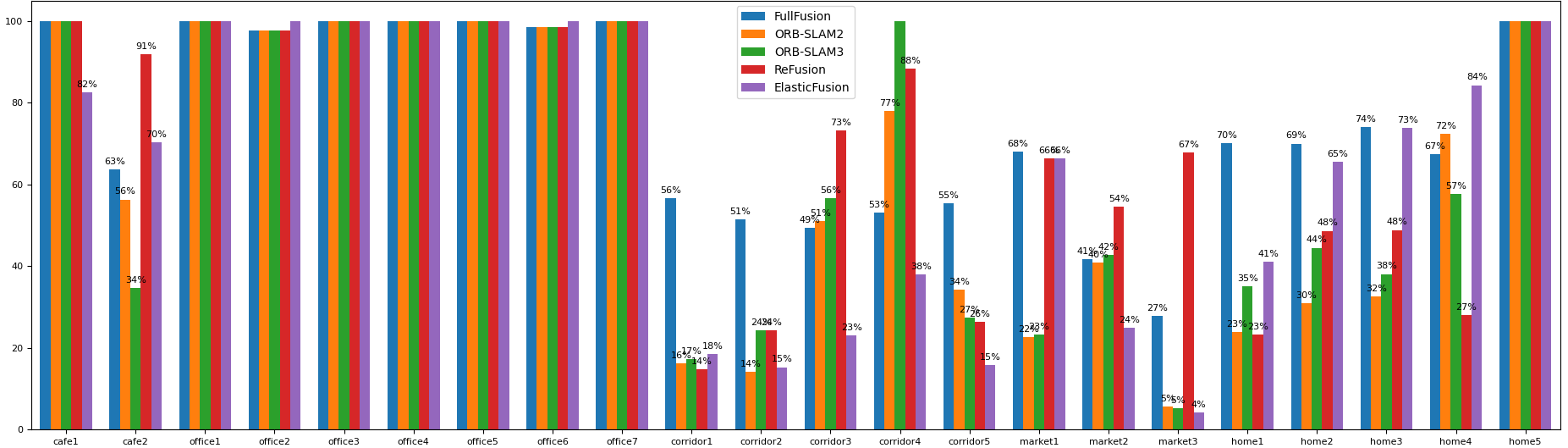}
      \caption{\textit{Correct Rate of Tracking} displayed as percentage of frames within an absolute error threshold on OpenLORIS sequences. The threshold values suggested in the dataset: $MAX\_ATE=1m$ for \textit{office}, $3m$ for \textit{home} and \textit{cafe}, and $5m$ for \textit{corridor} and \textit{market} sequences.}
  \label{fig:openloris}
  \vspace{-1em}
\end{figure*}

OpenLORIS is the most challenging of the datasets. Figure \ref{fig:openloris_ATE} illustrates ATE for a subset of the sequences evaluated. Figure \ref{fig:openloris} illustrates the CRT metric for all the sequences. Thus we can observe, for example,  that for the sequence cafe2 although the ATE may be less than 1 meter for ReFusion, ElasticFusion and FullFusion, their CRT illustrates significant portions of frames where the error was larger than 3 meters. ORB-SLAM2 and ORB-SLAM3 are severely affected in textureless environments. In particular, most of the \textit{corridor} and \textit{home} sequences disproportionately affect sparse algorithms.
ReFusion performed well in the presence of dynamic objects as long as they moved in a consistent fashion. However on the \textit{market} sequences, where persons often moved and stopped, artefacts were produced in the reconstruction, impacting the pose estimation accuracy. FullFusion performs well when it is able to recognise dynamic objects, but tends to drift whenever unknown objects enter the scene. 

\subsection{Other Observations}
In assessing the robustness of a SLAM system, one should consider not only variation across perturbations, but also matters of portability, setup, ease of use, consistency, and operation in previously untested environments.
\begin{table}[tb!]
\centering
\resizebox{\columnwidth}{!}{%
\begin{tabular}{|c|c|c|c|c|c|c|}
\hline
            & EF            & FULL       & ReF      & OVINS    & OS2       & OS3       \\ \hline
Jetson      & 40            & 25         & 0.2      & 10       & 5         & 5         \\
Laptop      & 50            & 30         & 10       & 17.5     & 5         & 10        \\
Workstation & 50            & 150        & 15       & 20       & 10        & 12        \\ \hline
\end{tabular}%
}
\caption{Average frame rate for each SLAM algorithm.}
\label{tab:framerate}
\vspace{-3em}
\end{table}

\noindent \textbf{Setup and execution} ---
ORB-SLAM2 and ORB-SLAM3 produced hard crashes (segfault) more than 10\% of the time across all platforms, requiring frequent restarts. Additionally, their reliance on old dependencies made it hard to identify working versions across all algorithms. Discrepancies in performance across platforms may relate to different versions of these dependencies. OpenVINS is highly sensitive to correct initial parameters, which may not always be available in deployment. We were not able to find working hyperparameters for the OpenLORIS dataset. FullFusion attempts to compute dynamic masks whether or not there are any dynamic objects in the scene, resulting in slightly lower accuracy as well as up to 80\% lower frame rate on compute-constrained platforms compared to disabling masks on sequences known to be static. ReFusion sees a drastic drop in frame rate to 0.2 FPS on the \textit{Jetson} from 10-15 FPS on \textit{Laptop} and \textit{Workstation}.

\begin{figure}[htbp]
  \centering
      \includegraphics[width=\columnwidth]{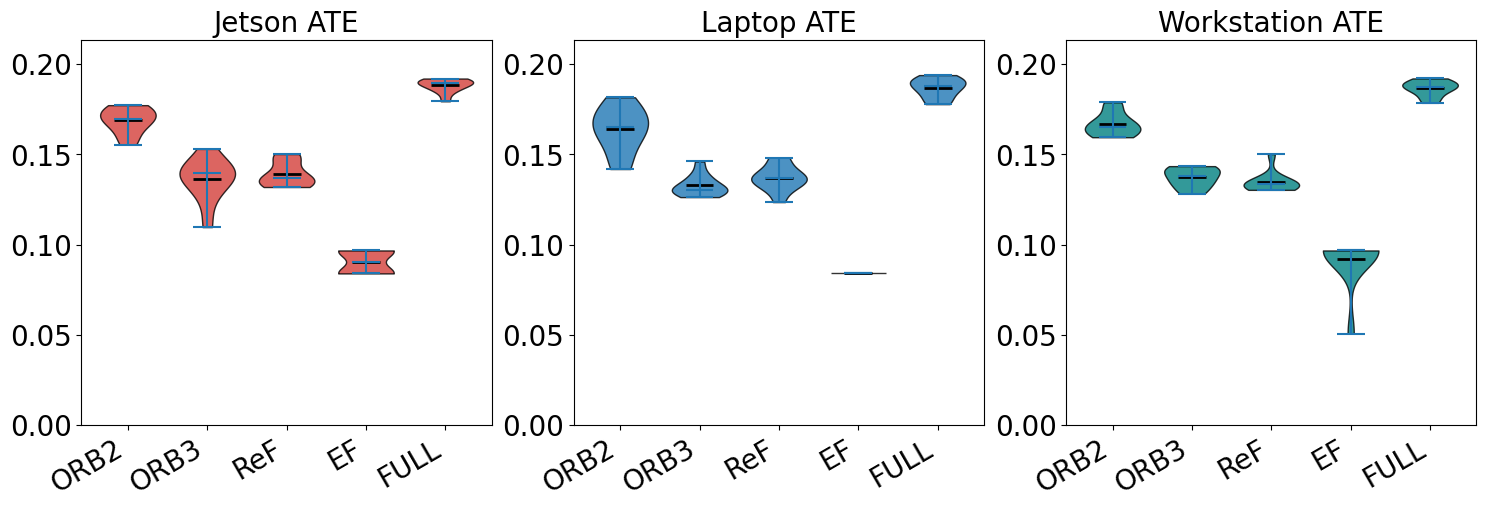}
      \caption{Consistency evaluation  using 30 runs on (\textit{fr1\_360}).}
  \label{fig:consistency}
  \vspace{-1em}
\end{figure}

\noindent \textbf{Consistency} --- While overall we have found no major discrepancies between the results on each platform, ORB-SLAM2 and ORB-SLAM3 were the least consistent across runs and across platforms, with some variability observed in other SLAM systems (see Figure \ref{fig:consistency}). At the other end, OpenVINS performed almost identically across all runs for any sequence on a given platform.

\subsection{Summary of Results} 

Table \ref{tab:overall} provides a summary of the results presented. The second to fourth columns show the dense SLAM systems. For the accuracy on baseline datasets (TUM, ICL-NUIM, EuRoC-MAV), no SLAM system is classified as \textit{Excellent} because, although their ATEs show good accuracy, they all have sequences where accuracy problems occur. ElasticFusion is fast and accurate when no perturbations are present but generally not robust due to the photometric error assuming fixed coefficients for the RGB channels. ReFusion is not very accurate on the baseline datasets and is sensitive to illumination changes due to the photometric error similar to ElasticFusion, however it is robust to dynamic objects. FullFusion is robust to illumination because it only uses depth data for mapping, but this can be a disadvantage in structureless areas. FullFusion has proven sensitive to unrecognized dynamic objects.

The fifth to seventh columns present the sparse SLAM systems. OpenVINS is the most consistent across runs and across platforms, and attains high accuracy on drone sequences, but is strictly visual-inertial and could not be tested on datasets without IMU data. ORB-SLAM2 and ORB-SLAM3 cover the broadest variety of input modalities. Consistent with the published ORB-SLAM3 \cite{ORBSLAM3_2020} results, but using different datasets, we have found that the addition of a VIO mode over ORB-SLAM2 and the multi-map merging scheme improves robustness against temporary tracking loss (usually caused by dynamic objects or fast movement). 

We use 5 categories to qualitatively describe the results. \textit{Excellent} means almost perfect - the system performs similarly with or without perturbations. This is only awarded in Illumination where FullFusion does not use color data and ORB-SLAM performs well even in low light. The category \textit{Very Good} covers the vast majority of perturbations and does not fail even when perturbations are significant. Nonetheless, a SLAM system may fail when encountering severe perturbations for a long period. If short term failures are encountered, it is often able to recover. The \textit{Good} category captures mostly robust outcomes. A good example is FullFusion which has robustness against a set number of classes, but may fail when encountering unknown classes. The category \textit{Acceptable} captures SLAM systems with some robustness, which can deal with perturbations for a short amount of time. For example, ORB-SLAM3 can recover well in the presence of dynamic objects, if they are encountered for a couple of frames or occupy only a small portion of the frame, but will fail otherwise.

\begin{table*}[tb!]
  \centering
      \includegraphics[width=0.9\textwidth]{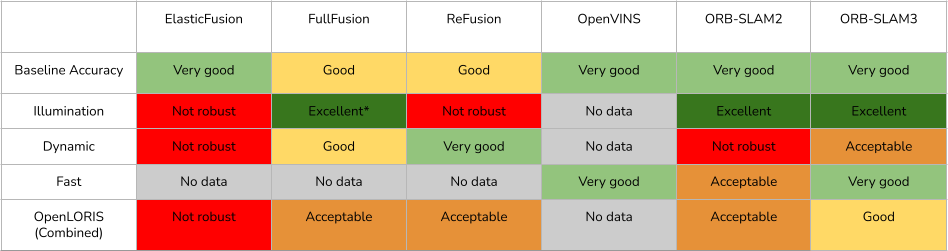}\\
      \footnotesize{*FullFusion is not impacted by illumination changes as it does not use color information.}
      \caption{Overall robustness -- A qualitative summary of the experiments.}
  \label{tab:overall}
\end{table*}

\section{Conclusions}
This paper  has presented  a  systematic  evaluation  of  the  robustness of 6 open-source state-of-the-art  SLAM  algorithms  with  respect  to  challenging conditions, such as fast motion, non-uniform illumination, and dynamic scenes. The experiments have covered 6 datasets across 3 computing platforms, in both  episodic and long-term operation settings. Thus, this evaluation is the most comprehensive study of the robustness of SLAM systems to date. By including the Nvidia Jetson Xavier platform, we also consider constraints associated with deployments on systems embedded within robots.

Overall, we have found that ORB-SLAM3 provides the best balance between baseline accuracy, illumination and fast changes, support for dynamic environments and Lifelong scenarios, although its FPS is below 15 (5 FPS on Jetson). Considering the three dense SLAM systems, FullFusion provides the best balance, but reaches 30 FPS only on the laptop and workstation (Jetson 25 FPS). ElasticFusion offers between 40-50 FPS processing on the three platforms, but its robustness falls below the other SLAM systems.

Finally, the sparse SLAM systems have proved more robust than the dense ones, probably because there are fewer data points which can negatively impact pose estimation. We consider that combining sparse tracking with dense 3D reconstruction will help systems build expressive representations while maintaining high robustness.

\section*{Acknowledgments}
This research is supported by the EPSRC, grant RAIN Hub EP/R026084/1. Mikel Luj\'an is supported by an Arm/RAEng Research Chair Award and a Royal Society Wolfson Fellowship. Thanks to Patrick Geneva for assisting with experiments on OpenVINS. Thanks to all researchers who provided the datasets.

\bibliographystyle{IEEEtran}
\bibliography{sample}
\end{document}